\def\year{2022}\relax
\documentclass[letterpaper]{article} 
\usepackage{aaai22}  
\usepackage{times}  
\usepackage{helvet}  
\usepackage{courier}  
\usepackage[hyphens]{url}  
\usepackage{graphicx} 
\usepackage{natbib}  
\usepackage{caption} 
\usepackage{algorithm}

\usepackage{newfloat}
\usepackage{listings}
\DeclareCaptionStyle{ruled}{labelfont=normalfont,labelsep=colon,strut=off} 
\floatstyle{ruled}
\newfloat{listing}{tb}{lst}{}
\floatname{listing}{Listing}

\usepackage{etoolbox}
\makeatletter
\AfterEndEnvironment{algorithm}{\let\@algcomment\relax}
\AtEndEnvironment{algorithm}{\kern2pt\hrule\relax\vskip3pt\@algcomment}
\let\@algcomment\relax
\newcommand\algcomment[1]{\def\@algcomment{\footnotesize#1}}
\renewcommand\fs@ruled{\def\@fs@cfont{\bfseries}\let\@fs@capt\floatc@ruled
  \def\@fs@pre{\hrule height.8pt depth0pt \kern2pt}%
  \def\@fs@post{}%
  \def\@fs@mid{\kern2pt\hrule\kern2pt}%
  \let\@fs@iftopcapt\iftrue}
\makeatother

\usepackage{graphicx}

\usepackage{tikz}
\usepackage{comment}
\usepackage{amsmath,amssymb} 
\usepackage{color}

\usepackage{subcaption}
\usepackage{booktabs}
\usepackage{multirow}
\usepackage{makecell}
\usepackage{wrapfig}
\usepackage{enumerate}
\usepackage{bbm}
\usepackage{algpseudocode}

\usepackage[accsupp]{axessibility}  

\newcommand{\eg}{\textit{e.g.}}
\newcommand{\ie}{\textit{i.e.}}

\title{{Darwinian} Model Upgrades: Model Evolving with Selective Compatibility}
\author{
Binjie Zhang$^{1,2}$\footnotemark[4] \qquad
Shupeng Su$^1$\footnotemark[4] \qquad
Yixiao Ge$^1$\footnotemark[5] \qquad
Xuyuan Xu$^3$ \\
Yexin Wang$^3$ \qquad
Chun Yuan$^4$ \qquad
Mike Zheng Shou$^2$ \qquad
Ying Shan$^1$ \\
{\small \footnotemark[4]~{equal contribution} \qquad
\footnotemark[5]~{corresponding author}} \\
}
\affiliations{
$^1$ARC Lab, Tencent PCG \qquad
$^2$National University of Singapore \\
$^3$AI Technology Center of Tencent Video \qquad
$^4$Tsinghua University \\
}

\begin{document}

\maketitle

\renewcommand{\thefootnote}{\fnsymbol{footnote}}

\begin{abstract}
The traditional model upgrading paradigm for retrieval requires recomputing all gallery embeddings before deploying the new model (dubbed as ``backfilling''), which is quite expensive and time-consuming considering billions of instances in industrial applications. BCT presents the first step towards backward-compatible model upgrades to get rid of backfilling. It is workable but leaves the new model in a dilemma between new feature discriminativeness and new-to-old compatibility due to the undifferentiated compatibility constraints. In this work, we propose Darwinian Model Upgrades (DMU), which disentangle the inheritance and variation in the model evolving with selective backward compatibility and forward adaptation, respectively. The old-to-new heritable knowledge is measured by old feature discriminativeness, and the gallery features, especially those of poor quality, are evolved in a lightweight manner to become more adaptive in the new latent space. We demonstrate the superiority of DMU through comprehensive experiments on large-scale landmark retrieval and face recognition benchmarks. DMU effectively alleviates new-to-new degradation 
and improves new-to-old compatibility, rendering a more proper model upgrading paradigm in large-scale retrieval systems.
\end{abstract}

\section{Introduction}

\begin{figure}[t]
\centering
\includegraphics[width=.85\linewidth]{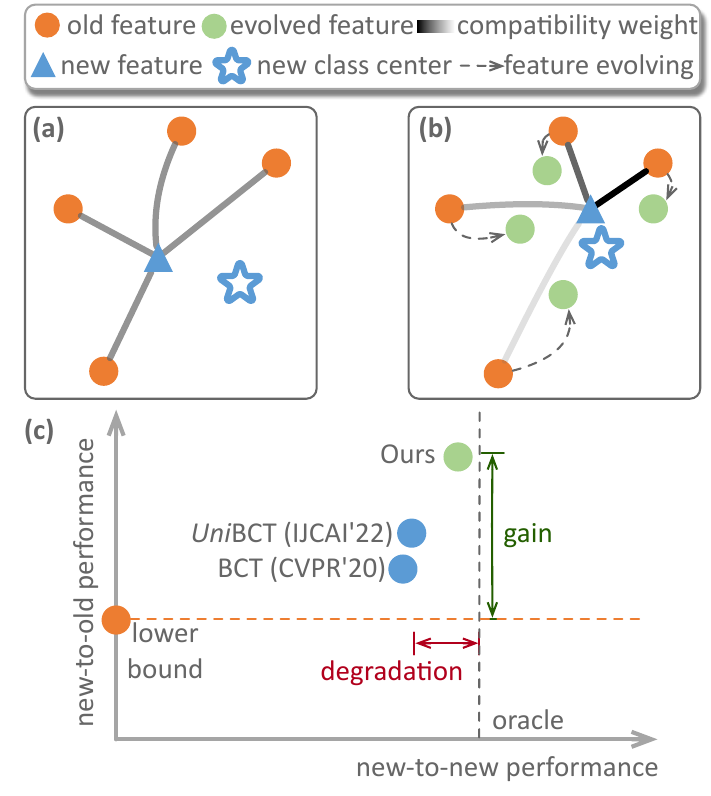}
\vspace{-8pt}
\caption{
\textbf{(a) Previous backward-compatible model upgrades:} the dilemma between new model capability and backward compatibility caused by undifferentiated inheritance. 
\textbf{(b) Darwinian model upgrades:} get out of the dilemma with selective backward compatibility/inheritance and forward adaptation/evolution. 
\textbf{(c) Experimental verification:} our DMU well alleviates the new-to-new retrieval degradation while boosting new-to-old compatibility on Google Landmark v2 dataset.}
\vspace{-8pt}
\label{fig:teaser}
\end{figure}

The paradigm of backward-compatible model upgrades was first introduced in \cite{shen2020cvpr}, and attracted increasing attention in the community
\cite{zhang2022iclr,meng2021learning,hu2022learning,zhang2022towards,su2022privacy}. Such a paradigm enables the backfill-free model upgrades\footnote{The term ``backfill'' indicates re-extracting all the gallery embeddings with the new model before its deployment.}, that is, the backward-compatible new models can be deployed on system immediately after training. It is especially beneficial for the large-scale retrieval systems in which the conventional offline backfilling may take weeks or even months for a model upgrade.

The essence of backward compatibility is to make old gallery embeddings and new query embeddings interoperable in the same latent space. Despite various efforts \cite{zhang2022towards,budnik2020asymmetric} to improve cross-model compatibility, a dilemma has yet to be solved that the new models have to sacrifice their internal capabilities (new-to-new) to maintain the compatibility with the older inferior models (new-to-old). 
Note that the new-to-new accuracy is also important for backward-compatible model upgrades as it affects the feature discriminativeness of the query and new gallery instances.
We argue that the dilemma is caused by the \textit{undifferentiated inheritance} in existing compatible methods for retrieval, where new embedding models are required to inherit both beneficial and harmful historical knowledge\footnote{In embedding applications (\textit{e.g.}, image retrieval), model knowledge is generally assessed by feature discriminativeness.}.

To this end, we present \textbf{D}arwinian \textbf{M}odel \textbf{U}pgrades\footnote{The name draws from an analogy to ``natural selection'' in Darwinism, but for model evolving.} (\textbf{DMU}), a new paradigm in which (i) the more discriminative old features that favor retrieval will tend to be inherited and (ii) the inferior old features will be adapted (upgraded) towards the new latent space. 
The former property well alleviates the new-to-new self-discrimination degradation which also improves the new-to-old retrieval with better new query features.
The latter property further enhances the new-to-old compatibility by evolving the old gallery features. 
DMU can properly relieve the dilemma between new model capability and compatibility.

Specifically, DMU is reached by the selective backward compatibility (inheritance) and the forward adaptation (variation). 
The backward-compatible regularization is re-weighted by the old feature discriminativeness, \textit{i.e.}, the probabilities of correctly identifying certain old features using the old classifier act as coefficients for the new-to-old compatible loss.
To further upgrade the old features especially those under quality,
a forward-adaptive module is trained to transform old gallery embeddings in a feature-to-feature fashion.
The forward-adaptive module is lightweight and requires marginal computational overhead.

We demonstrate the superiority of DMU against the other model upgrading methods on multiple widely-acknowledged image retrieval benchmarks, including the large-scale landmark retrieval and face recognition.
Following state-of-the-art methods \cite{gldv2-1rd:jeon20201st,gldv2-3rd:mei20203rd}, we perform model upgrades on GLDv2-train \cite{gldv2_2020} and MS1Mv3 \cite{deng2019arcface}, and evaluate retrieval on GLDv2-test, $\mathcal{R}$Oxford \cite{roxford_rparis_2018}, $\mathcal{R}$Paris, and IJB-C \cite{radenovic2018revisiting} respectively. 
DMU consistently surpasses the competitors in terms of both new-to-new and new-to-old retrieval performances.

Overall, we tackle the dilemma in existing backward-compatible model upgrades that new models have to sacrifice their feature discriminativeness to preserve the compatibility with old models by introducing Darwinian Model Upgrading paradigm with selective backward compatibility and forward adaptation. We extensively verify the effectiveness of our method in terms of both new-to-new and new-to-old retrieval on multiple public benchmarks. We hope our work would inspire the community to think out of the box and discover more proper model upgrading methods.

\section{Related Work}
\subsubsection{Backward-compatible model upgrades.}
BCT~\cite{shen2020cvpr} is the first work to introduce backward-compatible model upgrades paradigm by adding an influence loss when training the new model. It enforces the new features to approach the corresponding old class centroids, rendering the comparability between the new and old embedding spaces and bypassing the gallery backfilling. 
RACT~\cite{zhang2022iclr} takes a further step by introducing the hot-refresh backward-compatible model upgrades that backfills the gallery on-the-fly after the deployment of the backward-compatible new model.
Under this BCT framework, several work~\cite{budnik2020asymmetric,meng2021learning,zhang2022towards} makes efforts on designing the different training objectives to improve the backward-compatible retrieval performance.
Specifically, AML~\cite{budnik2020asymmetric} studies the adaptation of the common metric learning into asymmetric retrieval constraints when training the new model towards the old one, and finally introduces a better variant combining the contrastive and regression losses.
LCE~\cite{meng2021learning} proposes an alignment loss to align the new classifier with the old one and a tighter boundary loss for the new features towards the old class centers.
UniBCT~\cite{zhang2022towards} introduces a universal backward-compatible loss which refines the noisy old features with graph transition at first and then performs the backward-compatible training.
Besides studying the loss functions, CMP-NAS~\cite{duggal2021compatibility} further resorts to the neural architecture search technique to automatically search for the optimal compatibility-aware new neural framework for a given model.
Although making progress, the previous literature has to sacrifice the discriminativeness of the new model heavily as they put the same weight on the backward-compatible constraint towards each old embedding. It is sensitive to the outliers and accordingly we propose an adaptive re-weight scheme to alleviate this.

\subsubsection{Cross-model compatible transformation.}
Compatible feature transformation~\cite{chen2019r3,wang2020unified,ramanujan2022forward} also targets the compatible retrieval among different embedding models. 
However, it concerns training transformation modules to map the features from different models to a common space.
Specifically, \cite{chen2019r3} first introduces the embedding transformation problem in the face recognition occasion and tackle it with a R$^{3}$AN framework which transforms the source embeddings into the target embeddings as a regression task. CMC~\cite{wang2020unified} generalizes the problem in unlimited retrieval occasions and proposes the RBT (Residual Bottleneck Transformation) module with a training scheme combining classification loss, regression loss and KL-divergence loss.
FCT~\cite{ramanujan2022forward} further introduces the side information feature for each sample which facilitates the upgrade of old embeddings towards the unknown future new embeddings.
Although a feasible model upgrade paradigm, the mere cross-model transformation will lose the merits of the immediate deployment of new embedding model and upgrading the gallery features on-the-fly when compared to the backward-compatible paradigm above. Moreover, training transformation alone hardly achieves optimal performance as the feature upgrade module is generally lightweight and difficult to update the old features to a completely incompatible new space.

\begin{figure*}[t]
\centering
\includegraphics[width=1.\linewidth]{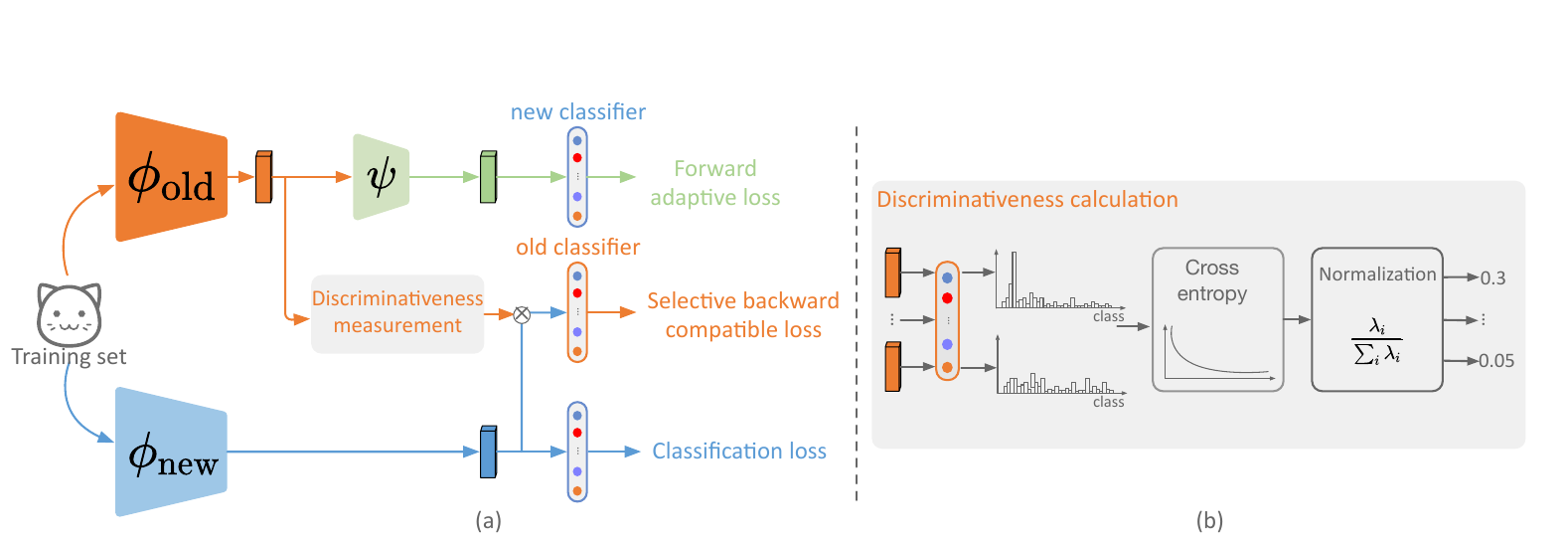}
\vspace{-16pt}
\caption{(a) The proposed {Darwinian Model Upgrades} framework which performs inheritance with selective compatibility and variation with forward adaptation.
(b) The details of discriminativeness measurement.}
\vspace{-8pt}
\label{fig:framework}
\end{figure*}

\subsubsection{Focal distillation.}
PCT~\cite{yan2020positive} proposes a focal distillation loss to relieve the ``negative flips" problem when training the new model which gives more distillation loss weights on the samples that are classified correctly by the old model. Considering both are re-weight scheme, our selective backward-compatible loss is different in: (1) PCT targets the classification task while our DMU focuses on the retrieval and constrains the feature compatibility instead of the classification probability similarity; (2) our re-weight scheme is adaptive to the discriminativeness of old features while the PCT one is only an indication function of whether correctly classified by the old model. 
We will demonstrate that DMU surpasses PCT distinctly in compatible retrieval.

\section{Methodology}

We study the task of efficient and effective model upgrades in large-scale embedding applications and take the widely-used image retrieval as an example, introduced in Sec. \ref{sec:prob}.
\cite{shen2020cvpr} breaks the traditional way of model upgrades and comes up with a new paradigm, backward-compatible model upgrades, that enables immediate deployment of the new model without the time-consuming backfilling.
Despite a feasible solution, it suffers from the undifferentiated inheritance, discussed in Sec. \ref{sec:bct}.
Consequently, we introduce a new paradigm, namely, Darwinian Model Upgrades, with selective backward compatibility and forward adaptation (see Sec. \ref{sec:dmu}). 
We elaborate each part as follows.

\subsection{Image Retrieval}\label{sec:prob}
Let the gallery be denoted as $\mathcal{G}$ and query as $\mathcal{Q}$. 
Given an embedding model $\phi: x \mapsto \mathbb{R}^{D}$, 
the task of image retrieval is to identify the object/content of interest in gallery resorting to the query-gallery similarities, \textit{i.e.}, $\|\phi(\mathcal{Q})-\phi(\mathcal{G})\|$.
Following state-of-the-art metric learning methods~\cite{gldv2-1rd:jeon20201st,gldv2-3rd:mei20203rd} in image retrieval, we employ classification as a pretext task in the form of ArcFace loss~\cite{deng2019arcface}.
Given an image $x$ of label $y$, the loss is formulated as
\begin{equation}\label{eq:arcface_loss}
  \begin{split}
      &\ell_{\rm arc}(x;\phi,\omega)= \\
      &- \log{ \frac{e^{s\cdot k(\phi(x),\omega(y), m)}}{e^{s\cdot k(\phi(x),\omega(y), m)}+\sum_{j\neq y}e^{s\cdot k(\phi(x),\omega(j), 0)} }}, 
  \end{split}
\end{equation}
where $m$ is the margin, $s$ is a scalar, and $\omega$ is a classifier with fully-connected layers. 
The kernel function is defined as $k(\phi(x),\omega(j), m)=\cos(\arccos\langle\phi(x),\omega(j)\rangle+m)$, where $\langle\cdot,\cdot\rangle$ represents the inner product.

\subsection{Backfill-free Model Upgrades: A Revisit}\label{sec:bct}

\subsubsection{Backward Compatible Training (BCT).}
BCT was first introduced by \cite{shen2020cvpr} to make the new and old features interchangeable in the same latent space. The training objectives are formulated as,
\begin{align}
    \begin{split}
        \mathop{\arg\min}_{\phi_\text{n}}~ ({\mathcal{L}_\text{new}} + {\mathcal{L}_\text{BC}}),
    \end{split}
\end{align}
where $\phi_{\rm n}$ is the new embedding model, $\mathcal{L}_{\rm new}$ is the self-discriminative loss for the new model, and $\ell_{\rm BC}$ is the backward-compatible constraint.

In a mini-batch $\mathcal{B}$, the self-discriminative loss for the new model is formulated as an ArcFace loss, such as
\begin{equation}\label{eq:base_loss}
  \begin{split}
      \mathcal{L}_{\rm new}= \frac{1}{|{\cal B}|} \sum_{x\in {\cal B}}  \ell_{\rm arc}(x;\phi_{\rm n},\omega_{\rm n}),
  \end{split}
\end{equation}
where $\omega_{\rm n}$ indicates the new classifier.
To make new features and old ones directly comparable with each other, 
backward compatibility is achieved by enforcing the new features to approach their corresponding old class centers,
\begin{equation}\label{eq:bct_loss}
  \begin{split}
      \mathcal{L}_{\rm BC}= \frac{1}{|{\cal B}|} \sum_{x\in {\cal B}}  \ell_{\rm arc}(x;\phi_{\rm n},\omega_{\rm o}),
  \end{split}
\end{equation}
where $\omega_{\rm o}$ is the old and fixed classifier.

\subsubsection{Backfill-free Model Upgrades.}
Given the interoperable old and new features, backfill-free model upgrades can be achieved where the new embedding model can be immediately deployed without re-extracting the gallery features. 
The retrieval accuracy of the new system is improved by more discriminative query features, \textit{i.e.},
\begin{equation}
\label{eq:ppmu-inequality}
    {\cal M}(\phi_{\rm o}(\mathcal{Q}), \phi_{\rm o}(\mathcal{G})) <
    {\cal M}(\phi_{\rm n}(\mathcal{Q}), \phi_{\rm o}(\mathcal{G})),
\end{equation}
where ${\cal M}(\cdot,\cdot)$ denotes the evaluation metric for retrieval (\textit{e.g.}, mAP used in landmark retrieval or TAR@FAR used in face recognition).
For simplicity, we will omit $\mathcal{Q}, \mathcal{G}$ and abbreviate  ${\cal M}(\phi_{\rm o}(\mathcal{Q}), \phi_{\rm o}(\mathcal{G}))$ as ${\cal M}(\phi_{\rm o}, \phi_{\rm o})$.
We denote the system-level model upgrading gain ($\Delta_{\uparrow}$) as
\begin{equation}
\label{eq:upgrade-gain}
    \Delta_{\uparrow} = \frac{\mathcal{M}(\phi_{\rm n}, \phi_{\rm o})-\mathcal{M}(\phi_{\rm o}, \phi_{\rm o})}{\mathcal{M}(\phi_{\rm o}, \phi_{\rm o})}.
\end{equation}

\begin{figure}[t]
\centering
\includegraphics[width=1.0\linewidth]{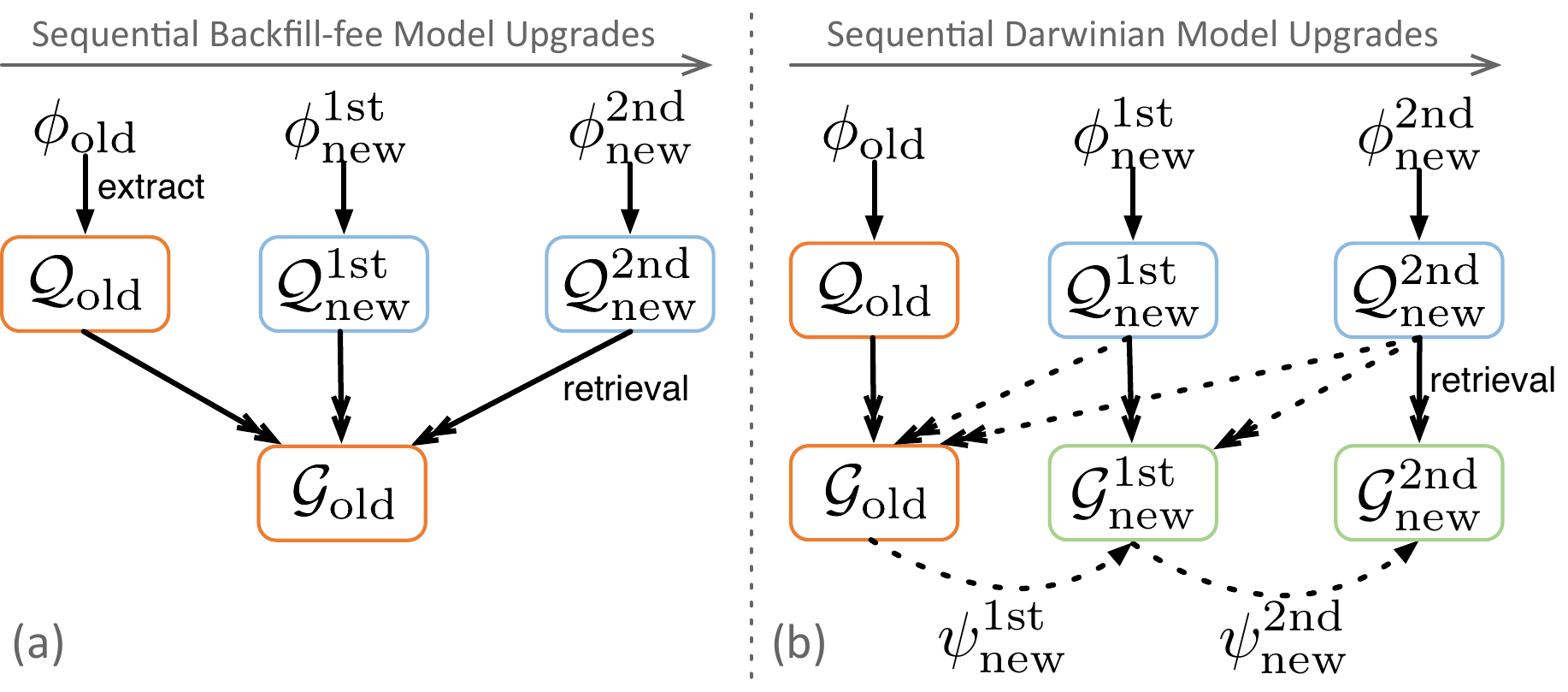}
\caption{Illustration of the backfill-free paradigm and Darwinian paradigm in sequential model upgrades.}
\label{fig:dmu}
\end{figure}

\subsubsection{Limitation.}
Despite the model upgrading gains achieved by new queries, we observe significant degradation of new-to-new retrieval accuracy compared to the model trained without backward compatibility (denoted as ``oracle''). It is actually due to the fact that new models have to sacrifice self-discriminativeness to maintain compatibility with the older and inferior models under the current \textit{undifferentiated inheritance} constraints, rendering BCT a sub-optimal solution for model upgrades. We evaluate the degree of such discriminativeness degradation ($\Delta_{\downarrow}$) with the following metric,
\begin{equation}
\label{eq:capacity-degradation}
    \Delta_{\downarrow} = \frac{\mathcal{M}(\phi_{\rm n}^{\rm oracle}, \phi_{\rm n}^{\rm oracle})-\mathcal{M}(\phi_{\rm n}, \phi_{\rm n})}{\mathcal{M}(\phi_{\rm n}^{\rm oracle}, \phi_{\rm n}^{\rm oracle})}.
\end{equation}

\subsection{Darwinian Model Upgrades (DMU)}\label{sec:dmu}

\subsubsection{Overview.}
Our DMU consists of a backward-compatible embedding model $\phi_{\rm n}$ and a lightweight forward-adaptive module $\psi$ (\eg, MLPs), as illustrated in Fig.~\ref{fig:framework}. 
To resolve the dilemma between upgrading gain ($\Delta_{\uparrow}$) and discriminativeness degradation ($\Delta_{\downarrow}$), DMU inherits the discriminative old knowledge with selective backward-compatible training and in the meanwhile evolves old embeddings towards new space with forward-adaptive training.
The overall training objectives of DMU can be formulated as,
\begin{align}
    \begin{split}
        \mathop{\arg\min}_{\phi_\text{n}, \psi}~ ({\mathcal{L}_\text{new} + \mathcal{L}_\text{SBC} + \mathcal{L}_\text{FA}}),
    \end{split}
\end{align}
where $\mathcal{L}_\text{SBC}$ is the selective backward-compatible loss, $\mathcal{L}_\text{FA}$ is the forward-adaptive loss, and $\mathcal{L}_\text{new}$ is the same as Equation \ref{eq:base_loss}.

\subsubsection{Inheritance with Selective Backward Compatibility.}

The key factor causing the discriminativeness degradation lies in the undifferentiated compatibility constraint where the new model is required to inherit both beneficial and harmful historical knowledge. We tackle the challenge with re-weighted backward-compatible objectives based on the old feature discriminativeness. We measure the feature discriminativeness with uncertainty metric from the information entropy theory,
\begin{equation}\label{eq:discriminativeness}
  \begin{split}
      &\Lambda(x;\phi,\omega) = \sum_{i\in |\mathcal{C}|}-p_i(x)\log p_i(x),~~~~\text{where}\\
      &p_i(x) = \frac{e^{<\phi(x),\omega(i)>}}{\sum_j e^{<\phi(x),\omega(j)>}}, 
      ~~(\sum_{i\in |\mathcal{C}|}p_i(x) = 1),
  \end{split}
\end{equation}
where $|\mathcal{C}|$ is the number of classes and $\omega$ is a classifier. $\Lambda(x;\phi,\omega)$ is inversely proportional to the feature discriminativeness.

For alleviating the negative effects of poor historical knowledge, we should punish the inheritance of samples with poor discriminativeness while stressing the good ones. Specifically, we use the old classifier $\omega_{\rm o}$ to evaluate the discriminativeness of old features, with which we formulate our selective backward-compatible loss as follows,
\begin{align}\label{eq:sbc_loss}
      &\mathcal{L}_{\rm SBC} = \sum_{x\in {\cal B}}\lambda(x) ~\ell_{\rm arc}(x;\phi_{\rm n},\omega_{\rm o}), ~~~~\hfill \text{where} \\
    &\lambda(x) = \frac{1-\text{softmax}[\Lambda(x;\phi_{\rm o},\omega_{\rm o})]}{|\mathcal{B}|-1},~~(\sum_{x\in \mathcal{B}}\lambda(x) = 1),\nonumber
\end{align}
where the softmax function is operated on a batch samples $(x_{1},..,x_{|\cal B|})$.
The old features with better discriminativeness will be allocated with larger backward-compatible weights.

\begin{figure}[t]
\centering
\includegraphics[width=0.7\linewidth]{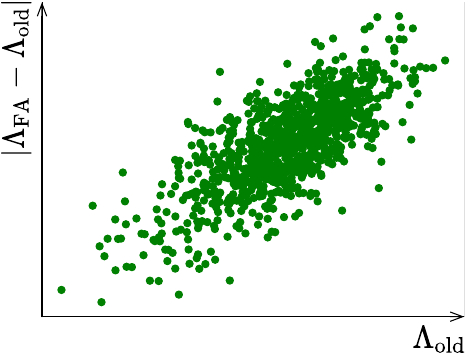}
\caption{Variation of entropy-based discriminativeness. We randomly select 1k gallery images from GLDv2-test and extract features with $\phi_{\rm old},\psi$. The horizontal axes is defined as $\Lambda_{\text{old}}=\Lambda(\mathcal{G};\phi_{\text{old}},\omega_\text{old})$, and the vertical axes as $\Lambda_{\text{FA}}=\Lambda(\mathcal{G};\psi(\phi_{\text{old}}),\omega_\text{new})$}
\label{fig:discriminativeness}
\end{figure}

\subsubsection{Variation with Forward Adaptation.}
The feature forward-adaptation module $\psi$ can be a simple Multi-Layer Perceptron (MLP), \ie, stack of fully-connected and non-linear activation layers (the detailed structure is listed in Experiments section).
$\psi$ upgrades the old features towards the new space with the following objective,
\begin{equation} \label{eq:fa-loss}
  \mathcal{L}_{\rm FA} =  \frac{1}{|\mathcal{B}|} \sum_{x\in {\cal B}}\ell_{\rm arc}(x; \psi(\phi_{\rm o}), \omega_{\rm n}).
\end{equation}
We do not specifically design a selective module for the forward adaptation, however, since the new model already inherits good old knowledge, it automatically selects poorer features for evolution.
As a demonstration, we exhibit in Fig.~\ref{fig:discriminativeness} that the old features with poor discriminativeness receive a large adaption degree than the good ones which is exactly attributed to our proposed selective backward-compatible constraints.

\begin{table*}[t]
\renewcommand\arraystretch{1.3}
\centering
\setlength{\tabcolsep}{0.1mm}{	
\begin{tabular}{llcccccccccccc}
	\toprule
	\multirow{2}[3]{*}{Scenario} & \multirow{2}[3]{*}{Method} & \multicolumn{4}{c}{GLDv2-test} & \multicolumn{4}{c}{$\mathcal{R}$Oxford} &
	\multicolumn{4}{c}{$\mathcal{R}$Paris} \\
	\cmidrule(r){3-6}
	\cmidrule(r){7-10}
	\cmidrule(r){11-14}
	~ & ~ & ${\cal M}(\phi_{\rm n},\phi_{\rm n})$ & ${\cal M}(\phi_{\rm n},\phi_{\rm o})$ & \quad\textcolor[RGB]{0,200,0}{$\Delta_{\uparrow}$}\qquad & \quad\textcolor[RGB]{200,0,0}{$\Delta_{\downarrow}$}\quad
	& ${\cal M}(\phi_{\rm n},\phi_{\rm n})$ & ${\cal M}(\phi_{\rm n},\phi_{\rm o})$ & \quad\textcolor[RGB]{0,200,0}{$\Delta_{\uparrow}$}\qquad & \quad\textcolor[RGB]{200,0,0}{$\Delta_{\downarrow}$}\quad
	& ${\cal M}(\phi_{\rm n},\phi_{\rm n})$ & ${\cal M}(\phi_{\rm n},\phi_{\rm o})$ & \quad\textcolor[RGB]{0,200,0}{$\Delta_{\uparrow}$}\qquad & \quad\textcolor[RGB]{200,0,0}{$\Delta_{\downarrow}$}\quad \\
	\midrule

	\multirow{4}{*}{\makecell[l]{30\%data\\$\rightarrow$ \\ 100\%data}} &
	$\phi_{\rm o}^{r50}$ &  18.93 &  - &  - & -  &  64.33 &  - &  - &  - & 82.94  & -  &  - & - \\
	
	~ & $\phi_{\rm n}^{r50}$ (oracle)& 25.58  &  - &  - & -  & 75.88  & -  & -  & -  &  86.09  & - & - &-   \\
	
	~ & $\phi_{\rm n}^{r50}$ (BCT) &  24.55 & 21.12  &  \textcolor[RGB]{0,200,0}{11.57} & \textcolor[RGB]{200,0,0}{4.03}  &  73.74 & 64.75  & \textcolor[RGB]{0,200,0}{0.65}  & \textcolor[RGB]{200,0,0}{2.82}  & 85.73  & 83.53  & \textcolor[RGB]{0,200,0}{0.71}  & \textcolor[RGB]{200,0,0}{0.42}    \\
	
	~ & $\phi_{\rm n}^{r50}$ (ours) &  \textbf{25.00} & \textbf{22.56}  & \textcolor[RGB]{0,200,0}{\textbf{19.18}}  & \textcolor[RGB]{200,0,0}{\textbf{2.27}}  &  \textbf{74.07} & \textbf{66.68}  & \textcolor[RGB]{0,200,0}{\textbf{3.65}}  & \textcolor[RGB]{200,0,0}{\textbf{2.39}}  & \textbf{86.59}  &  \textbf{84.48} & \textcolor[RGB]{0,200,0}{\textbf{1.86}}  & \textcolor[RGB]{200,0,0}{\textbf{-0.58}}  \\
	\hline

	\multirow{4}{*}{\makecell[l]{30\%data\\$\rightarrow$ \\ 70\%data}} &
	$\phi_{\rm o}^{r50}$ &  18.93 &  - &  - & -  &  64.33 & -  & -  &  - & 82.94  & -  & - &-  \\
	
	~ & $\phi_{\rm n}^{r50}$ & 23.66  &  - &  - & -  &  72.20 & -  & -  &  - &  86.64 &  - & - & -   \\
	
	~ & $\phi_{\rm n}^{r50}$ (BCT) &  23.47 & 20.74  & \textcolor[RGB]{0,200,0}{9.56}  & \textcolor[RGB]{200,0,0}{0.80}   &  72.91 & 63.73  & \textcolor[RGB]{0,200,0}{-0.93}  & \textcolor[RGB]{200,0,0}{-0.98}   &  \textbf{86.44} &  83.4 & \textcolor[RGB]{0,200,0}{0.55}  & \textcolor[RGB]{200,0,0}{\textbf{0.23}}   \\
	
	~ & $\phi_{\rm n}^{r50}$ (ours) &  \textbf{23.76} & \textbf{21.70}  & \textcolor[RGB]{0,200,0}{\textbf{14.63}}  & \textcolor[RGB]{200,0,0}{\textbf{-0.42}}   & \textbf{73.01}  &  \textbf{64.14} & \textcolor[RGB]{0,200,0}{\textbf{-0.30}}  & \textcolor[RGB]{200,0,0}{\textbf{-1.12}}  &  85.90 & \textbf{84.64} &  \textcolor[RGB]{0,200,0}{\textbf{2.05}}  & \textcolor[RGB]{200,0,0}{0.85}  \\
	\hline

	\multirow{4}{*}{\makecell[l]{30\%class\\$\rightarrow$\\100\%class}} &
	$\phi_{\rm o}^{r50}$ &  17.53 & -  & -  & -  &  70.07 & -  & -  & -  & 83.62  &  - & - & -   \\
	
	~ & $\phi_{\rm n}^{r50}$ & 25.58  &  - & -  & -  & 75.88  & -  & -  & -  & 86.09  & -  & - & -    \\
	
	~ & $\phi_{\rm n}^{r50}$ (BCT) & 25.10  & 24.39  & \textcolor[RGB]{0,200,0}{39.13}  & \textcolor[RGB]{200,0,0}{1.88}  & 73.99  &  73.34 & \textcolor[RGB]{0,200,0}{4.67}  & \textcolor[RGB]{200,0,0}{2.49}  &  85.64 & 83.97  & \textcolor[RGB]{0,200,0}{0.42}  & \textcolor[RGB]{200,0,0}{0.52}   \\
	
	~ & $\phi_{\rm n}^{r50}$ (ours) & \textbf{25.45}  &  \textbf{25.01} & \textcolor[RGB]{0,200,0}{\textbf{42.67}}  & \textcolor[RGB]{200,0,0}{\textbf{0.51}}   &  \textbf{74.13} & \textbf{73.78} & \textcolor[RGB]{0,200,0}{\textbf{5.29}}  & \textcolor[RGB]{200,0,0}{\textbf{2.31}}  &  \textbf{86.64} & \textbf{84.52} & \textcolor[RGB]{0,200,0}{\textbf{1.08}}  & \textcolor[RGB]{200,0,0}{\textbf{-0.64}}   \\
	\hline

	\multirow{4}{*}{\makecell[l]{resnet50\\$\rightarrow$\\resnet101}} &
	$\phi_{\rm o}^{r50}$ &  18.93 &  - &  - & -  &  64.33 & -  & -  &  - & 82.94  & -  & - &- \\
	
	~ & $\phi_{\rm n}^{r101}$ &  27.42 & -  & -  & -  & 75.89  &  - &  - & -  &  86.62 &  - &  - &  - \\
	
	~ & $\phi_{\rm n}^{r101}$ (BCT) &  25.56 &  21.68 & \textcolor[RGB]{0,200,0}{14.53}  & \textcolor[RGB]{200,0,0}{6.78}  & 75.7  &  66.12 & \textcolor[RGB]{0,200,0}{2.78}  & \textcolor[RGB]{200,0,0}{0.25}  &  87.35 &  85.59 & \textcolor[RGB]{0,200,0}{3.20}  & \textcolor[RGB]{200,0,0}{-0.84}   \\
	
	~ & $\phi_{\rm n}^{r101}$ (ours) &  \textbf{25.92} & \textbf{22.38}  & \textcolor[RGB]{0,200,0}{\textbf{18.23}}  & \textcolor[RGB]{200,0,0}{\textbf{5.47}}   & \textbf{76.49}  & \textbf{66.35} & \textcolor[RGB]{0,200,0}{\textbf{3.14}}  & \textcolor[RGB]{200,0,0}{\textbf{-0.79}}  & \textbf{88.08}  & \textbf{85.93} & \textcolor[RGB]{0,200,0}{\textbf{3.61}}  & \textcolor[RGB]{200,0,0}{\textbf{-1.69}}   \\

	\bottomrule
\end{tabular}
}
\vspace{-8pt}
\caption{Comparison of baselines (BCT) and our method (DMU) in various compatible scenarios on Landmark Retrieval datasets (\ie, GLDv2-test, ROxord and RParis). All models are trained on GLDv2-train dataset. The evaluation protocols are mAP ($\mathcal{M}(\cdot,\cdot)$, larger is better), upgrade gain ($\Delta_{\uparrow}$, larger is better) and discriminativeness degradation ($\Delta_{\downarrow}$, smaller is better). The architectures are ResNet50 (r50) and ResNet101 (r101). A negative value for degradation means that the degradation has been fully resolved, and the new model even benefits from backward-compatible regularization.
}
\label{tab:exp-landmark-retrieval}
\vspace{-8pt}
\end{table*}

\subsubsection{Darwinian Model Upgrades.} 
As depicted in Fig.~\ref{fig:dmu}, our proposed new paradigm, Darwinian Model Upgrades, improves the query embedding quality by the selective backward-compatible new model in the meanwhile improves the gallery embeddings by the feature forward-adaptation head. 
We formulate the objective of DMU formally as follows,

\begin{align}
\label{eq:ppmu-inequality-2}
    {\cal M}(\phi_{\rm o}, \phi_{\rm o}) <
    {\cal M}(\phi_{\rm n}, \phi_{\rm o}) < 
    {\cal M}(\phi_{\rm n}, \psi \circ \phi_{\rm o}) ,
\end{align}
where $\phi_{\rm n},\psi$ are the selective backward-compatible new model and feature forward-adaptation module. 
As will be shown later, performing DMU can bring consistent improvements for the retrieval performance and especially benefits sequential upgrades as it relieves the shackle of the new model for keeping compatibility to the oldest gallery.

\begin{table}[t]
\renewcommand\arraystretch{1.3}
\centering
\setlength{\tabcolsep}{0.1mm}{	
\begin{tabular}{llcccc}
	\toprule
	\multirow{2}[3]{*}{Scenario} & \multirow{2}[3]{*}{Method} & \multicolumn{4}{c}{IJB-C verification} \\
	\cmidrule(r){3-6}
	~ & ~ & ${\cal M}(\phi_{\rm n},\phi_{\rm n})$ & ${\cal M}(\phi_{\rm n},\phi_{\rm o})$ & \quad\textcolor[RGB]{0,200,0}{$\Delta_{\uparrow}$}\quad & \quad\textcolor[RGB]{200,0,0}{$\Delta_{\downarrow}$}\quad \\
	\midrule

	\multirow{4}{*}{\makecell[l]{30\%data\\$\rightarrow$ \\ 100\%data}} &
	$\phi_{\rm o}^{ir18}$ &  91.67 &  - &  - & -  \\
	~ & $\phi_{\rm n}^{ir18}$ (oracle)& 93.89  &  - &  - & - \\
	~ & $\phi_{\rm n}^{ir18}$ (BCT) & 93.96 & 92.82  &  \textcolor[RGB]{0,200,0}{1.25}  & \textcolor[RGB]{200,0,0}{-0.07} \\
	~ & $\phi_{\rm n}^{r18}$ (ours) & \textbf{94.17} & \textbf{92.96 } &  \textcolor[RGB]{0,200,0}{\textbf{1.41}}  & \textcolor[RGB]{200,0,0}{\textbf{-0.30}} \\	
	\hline

	\multirow{4}{*}{\makecell[l]{30\%data\\$\rightarrow$ \\ 70\%data}} &
	$\phi_{\rm o}^{ir18}$ &  91.67 &  - &  - & -  \\
	~ & $\phi_{\rm n}^{ir18}$ (oracle)& 93.79  &  - &  - & - \\
	~ & $\phi_{\rm n}^{ir18}$ (BCT) & 93.69   & 92.61  & \textcolor[RGB]{0,200,0}{1.03}  & \textcolor[RGB]{200,0,0}{0.11}\\
	~ & $\phi_{\rm n}^{ir18}$ (ours) &  \textbf{93.96} & \textbf{92.77} &  \textcolor[RGB]{0,200,0}{\textbf{1.20}}  & \textcolor[RGB]{200,0,0}{\textbf{-0.18}}  \\

	\bottomrule
\end{tabular}
}
\label{tab:exp-face-recognition}
\vspace{-8pt}
\caption{Comparison of baselines (BCT, UniBCT) and our method (DMU) on the Face Recognition dataset (\ie, IJB-C). All models are trained on MS1M-v3 dataset. The evaluation protocols are TAR@FAR=$1e^{-4}$ ($\mathcal{M}(\cdot,\cdot)$, larger is better), upgrade gain ($\Delta_{\uparrow}$, larger is better) and discriminativeness degradation ($\Delta_{\downarrow}$, smaller is better). The architecture is iResNet18 (ir18).}
\end{table}

\section{Experiments}
In this section, we will validate the proposed Darwinian Model Upgrades on the widely-acknowledged image retrieval datasets including the Google Landmark v2~\cite{gldv2_2020}, $\mathcal{R}$Oxford~\cite{roxford_rparis_2018}, $\mathcal{R}$Paris~\cite{roxford_rparis_2018}, MS1Mv3~\cite{deng2019arcface} and IJB-C~\cite{ijbc_maze2018iarpa}.

\subsection{Experiment settings}
\subsubsection{Datasets.} 
\textbf{(1) Landmark Retrieval:}
We adopt the GLDv2-train-clean version~\cite{gldv2_2020} as training set which is comprised of 1,580,470 images in 81,313 landmarks. 
After the training, the retrieval performance is evaluated on GLDv2-test~\cite{gldv2_2020}, Revisited Oxford ($\mathcal{R}$Oxford)~\cite{roxford_rparis_2018}, and Revisited Paris ($\mathcal{R}$Paris)~\cite{roxford_rparis_2018}. 
For GLDv2-test, there are 750 query images and 761,757 gallery images. As to $\mathcal{R}$Oxford, it contains 70 query images and 4,993 gallery images. $\mathcal{R}$Paris is comprised of 70 query images and 6,322 gallery images.
\textbf{(2) Face Recognition:} 
All models are trained on MS1Mv3~\cite{deng2019arcface} dataset, which contains 5,179,510 training images with 93,431 labels.
To evaluate the model generalization, we conduct face verification on {IJB-C}~\cite{ijbc_maze2018iarpa} test set, which provides 469,376 templates pairs.

\subsubsection{Metric.}
\textbf{(1) Landmark Retrieval:}
We adopt the cosine similarity for ranking and the retrieval results are measured with the mean Average Precision (mAP). 
Specifically, following the test protocols proposed in \cite{gldv2_2020,roxford_rparis_2018}, we use mAP@100 for GLDv2-test, and mAP@all in the \textbf{Medium} evaluation setup (\textit{Easy} and \textit{Hard} images are treated as positive, while \textit{Unclear} are ignored) for $\mathcal{R}$Oxford and $\mathcal{R}$Paris.
\textbf{(2) Face Recognition:} 
We calculate the true acceptance rate (TAR) at different false acceptance rates (FAR) for template pairs. In cross-model retrieval, we extract the first template with the new model, and the second with the old model.

\begin{table}[t]
\renewcommand\arraystretch{1.3}
\centering
\setlength{\tabcolsep}{0.1mm}{	
\begin{tabular}{lcccc}
	\toprule
	\multirow{2}[3]{*}{$\Lambda$ type} & \multicolumn{4}{c}{GLDv2-test} \\
	\cmidrule(r){2-5}
	~ & ${\cal M}(\phi_{\rm n},\phi_{\rm n})$ & ${\cal M}(\phi_{\rm n},\phi_{\rm o})$ & \quad\textcolor[RGB]{0,200,0}{$\Delta_{\uparrow}$}\quad & \quad\textcolor[RGB]{200,0,0}{$\Delta_{\downarrow}$}\quad \\
	\midrule
    uniform (BCT) & 24.55 & 21.12 &  \textcolor[RGB]{0,200,0}{11.57}  & \textcolor[RGB]{200,0,0}{4.03}  \\
	least conf.& 24.88  &  22.25 &  \textcolor[RGB]{0,200,0}{17.54}  & \textcolor[RGB]{200,0,0}{2.74} \\
	entropy &  \textbf{25.00} &  \textbf{22.56} &  \textcolor[RGB]{0,200,0}{19.18}  & \textcolor[RGB]{200,0,0}{2.27}  \\

	\bottomrule
\end{tabular}
}
\vspace{-8pt}
\caption{Comparison of different discriminativeness measurement in  30\%data$\rightarrow$100\%data on GLDv2-test. The backbone is ResNet50.}
\label{tab:exp-discriminativeness}
\vspace{-8pt}
\end{table}

\begin{table}[t]
\renewcommand\arraystretch{1.3}
\centering
\setlength{\tabcolsep}{0.1mm}{	
\begin{tabular}{lcccccc}
	\toprule
	\multirow{2}[3]{*}{Method} & \multirow{2}[3]{*}{$\mathcal{L}_{\rm SBC}$}  & \multirow{2}[3]{*}{$\mathcal{L}_{\rm FA}$} & \multicolumn{4}{c}{GLDv2-test} \\
	\cmidrule(r){4-7}
	~ & ~ & ~ & ${\cal M}(\phi_{\rm n},\phi_{\rm n})$ & ${\cal M}(\phi_{\rm n},\phi_{\rm o})$ & \quad\textcolor[RGB]{0,200,0}{$\Delta_{\uparrow}$}\quad & \quad\textcolor[RGB]{200,0,0}{$\Delta_{\downarrow}$}\quad \\
	\midrule

	$\phi_{\rm n}^{r50}$(BCT) & - & - & 24.55 & 21.12& \textcolor[RGB]{0,200,0}{11.57}  & \textcolor[RGB]{200,0,0}{4.03}   \\
	$\phi_{\rm n}^{r50}$ (ours)& \checkmark & - & 24.75 &  21.60 &  \textcolor[RGB]{0,200,0}{14.10}  & \textcolor[RGB]{200,0,0}{3.24} \\
	$\phi_{\rm n}^{r50}$ (ours) & - & \checkmark &  24.72  & 22.42  & \textcolor[RGB]{0,200,0}{18.44}  & \textcolor[RGB]{200,0,0}{3.36} \\
	$\phi_{\rm n}^{r50}$ (ours) & \checkmark & \checkmark &   \textbf{25.00} &\textbf{ 22.56} & \textcolor[RGB]{0,200,0}{\textbf{19.18}}  & \textcolor[RGB]{200,0,0}{\textbf{2.27}}  \\
	\bottomrule
\end{tabular}
}
\vspace{-8pt}
\caption{Ablation of our Darwinian Model Upgrades paradigm in 30\%data$\rightarrow$100\%data on GLDv2-test. The backbone is ResNet50.}
\label{tab:exp-ablation}
\end{table}

\subsubsection{Compatible Scenarios.} 
To simulate the different model upgrade occasions, we randomly split the training set with different percentages (30\%, 70\% and 100\%). For extended-data (30\%data $\rightarrow$ 100\%data) scenario, the old training set and the new one share all landmarks; for extended-class (30\%class $\rightarrow$ 100\%class) scenario, the old training set only covers part of all labels.
Besides, we verify our algorithm in different model architectures ( resnet50$\rightarrow$resnet101).
Details about data allocation are in supplemental materials.

\subsubsection{Training details.}
Most of our basic settings follows the experience in \cite{yuqi20212nd,ozaki2019large,henkel2020supporting}. Specifically, we adopt the ResNet~\cite{he2016deep} (resnet50 by default unless specified otherwise) as the backbone of embedding model and substitute the global average pooling layer by Generalized-Mean (GeM) pooling~\cite{radenovic2018fine} with hyper-parameter p=3. A fully-connected (fc) layer is appended in the end which transforms the pooling features into the output embedding (the dimension is set as 1024). 
For the feature upgrade module $\psi$, we adopt [fc-bn-relu] as basic block and stack 3 of this with a output fc layer as the final MLP construction. 
The input image is resized to $224\times 224$ for training and $224\times 224$ for inference. 
Random image augmentation is applied which includes the random resized cropping and horizontal flipping.
With 6 Tesla V100 for training, the batch size per GPU is set as 256 for the embedding model and 512 for the feature upgrade module.
SGD optimizer with 0.9 momentum and 10$^{-4}$ weight decay is adopted. Besides, we uniformly use the cosine lr scheduler with 1 warm-up epoch in the total running of 30 epochs. The initial learning rate is set as 0.1 . For the training objective, we set hyper-parameters in ArcFace loss as $s=30,~m=0.3$ in landmark retrieval, $s=64,~m=0.5$ in face recognition.

\begin{table}[t]
\renewcommand\arraystretch{1.3}
\centering
\setlength{\tabcolsep}{0.1mm}{	
\begin{tabular}{lcccc}
	\toprule
	\multirow{2}[3]{*}{Dimension} & \multicolumn{4}{c}{GLDv2-test} \\
	\cmidrule(r){2-5}
	~ & ${\cal M}(\phi_{\rm n},\phi_{\rm n})$ & ${\cal M}(\phi_{\rm n},\phi_{\rm o})$ & \quad\textcolor[RGB]{0,200,0}{$\Delta_{\uparrow}$}\quad & \quad\textcolor[RGB]{200,0,0}{$\Delta_{\downarrow}$}\quad \\
	\midrule
	
	512 &  24.78 &  22.12 &  \textcolor[RGB]{0,200,0}{16.85}  & \textcolor[RGB]{200,0,0}{3.13}  \\
	1024 & \textbf{25.00}  &  22.56 &  \textcolor[RGB]{0,200,0}{19.18}  & \textcolor[RGB]{200,0,0}{\textbf{2.27}} \\
	2048 & 24.96  &  \textbf{22.77} &  \textcolor[RGB]{0,200,0}{\textbf{20.29}}  & \textcolor[RGB]{200,0,0}{2.42} \\

	\bottomrule
\end{tabular}
}
\vspace{-8pt}
\caption{Comparison of different dimension in forward-apative head in  30\%data$\rightarrow$100\%data on GLDv2-test. The backbone is ResNet50.}
\label{tab:exp-fa-dimension}
\end{table}

\subsection{Analysis of DMU}

\subsubsection{Different model upgrade occasions with DMU.}
As shown in Table~\ref{tab:exp-landmark-retrieval}, our DMU surpasses baselines on various practical model upgrade occasions, which comprehensively demonstrates the effectiveness of our approach.

\subsubsection{Effectiveness of Selective Backward Compatibility.}
To demonstrate the proposed selective backward compatibility can alleviate the issue of discriminativeness degradation, we only train the backbone without the forward-adaptive head. Results in Table~\ref{tab:exp-ablation} show that inheriting good old knowledge will benefit the new model training.

Except from the entropy-based discriminativeness measurement, we verify another approach based on least confidence, defined as the following,
\begin{equation}\label{eq:discriminativeness-least-confidence}
  \begin{split}
      \Lambda(x;\phi,\omega) &= \sum_{i}\mathbbm{1}(i=y)\cdot p_i(x),\\
      \lambda(x) &=\text{softmax}(\Lambda(x;\phi,\omega)),
  \end{split}
\end{equation}
where $p_i(x)$ is calculated by the same formulation in Equation~\ref{eq:discriminativeness}. 
As illustrated in Table~\ref{tab:exp-discriminativeness}, different forms of discriminativeness measurement alleviate new-to-new degradation and boost new-to-old retrieval performance, which indicates the generalization of our method.

\begin{table*}[th]
\renewcommand\arraystretch{1.2}
\centering
\setlength{\tabcolsep}{1.8mm}{	
\begin{tabular}{cccccccccccc}
	\toprule
	\multirow{2}[3]{*}{Scenario} & \multirow{2}[3]{*}{Method} & \multicolumn{4}{c}{1st generation} & \multicolumn{4}{c}{2nd generation} \\
	\cmidrule(r){3-6}
	\cmidrule(r){7-10}
	~ & ~ & ${\cal M}(\phi_{\rm n},\phi_{\rm n})$ & ${\cal M}(\phi_{\rm n},\phi_{\rm o})$ & \quad\textcolor[RGB]{0,200,0}{$\Delta_{\uparrow}$}\quad & \quad\textcolor[RGB]{200,0,0}{$\Delta_{\downarrow}$}\quad &  ${\cal M}(\phi_{\rm n},\phi_{\rm n})$ & ${\cal M}(\phi_{\rm n},\phi_{\rm o})$ & \quad\textcolor[RGB]{0,200,0}{$\Delta_{\uparrow}$}\quad & \quad\textcolor[RGB]{200,0,0}{$\Delta_{\downarrow}$}\quad \\
	\midrule

	\multirow{3}{*}{\makecell[c]{Extended data:\\25\%$\rightarrow$50\%$\rightarrow$75\%}}
	& $\phi_{\rm new}^{\rm oracle}$/$\phi_{\rm old}$ & 22.39/17.89 & - & - & - & 24.04/17.89 & - & - & - \\
	& BCT  & 21.47 & 18.88  & \textcolor[RGB]{0,200,0}{5.53}  & \textcolor[RGB]{200,0,0}{4.11} & 23.52 & 19.82 & \textcolor[RGB]{0,200,0}{10.79}  & \textcolor[RGB]{200,0,0}{2.16} \\
	& ours & 22.66 & 19.84 & \textcolor[RGB]{0,200,0}{10.90}  & \textcolor[RGB]{200,0,0}{-1.21} & 23.88 & 21.16  & \textcolor[RGB]{0,200,0}{18.28}  & \textcolor[RGB]{200,0,0}{0.67} \\

	\bottomrule
\end{tabular}
}
\caption{Sequential model upgrades with DMU on GLDv2-test. The backbone is ResNet50.}
\label{tab:exp-sequential-upgrades}
\vspace{-8pt}
\end{table*}

\begin{table}[t]
\renewcommand\arraystretch{1.3}
\centering
\setlength{\tabcolsep}{0.1mm}{	
\begin{tabular}{lcccc}
	\toprule
	\multirow{2}[3]{*}{Comp. loss} & \multicolumn{4}{c}{GLDv2-test} \\
	\cmidrule(r){2-5}
	~ & ${\cal M}(\phi_{\rm n},\phi_{\rm n})$ & ${\cal M}(\phi_{\rm n},\phi_{\rm o})$ & \quad\textcolor[RGB]{0,200,0}{$\Delta_{\uparrow}$}\quad & \quad\textcolor[RGB]{200,0,0}{$\Delta_{\downarrow}$}\quad \\
	\midrule
    BCT (ArcFace)  & 24.55 & 21.12& \textcolor[RGB]{0,200,0}{11.57}  & \textcolor[RGB]{200,0,0}{4.03}  \\
	DMU (ArcFace)  & 25.00 &\textbf{ 22.56} & \textcolor[RGB]{0,200,0}{\textbf{19.18}}  & \textcolor[RGB]{200,0,0}{2.27}   \\
	BCT (Regression)  & 25.11 & 19.96 & \textcolor[RGB]{0,200,0}{5.44}  & \textcolor[RGB]{200,0,0}{1.84}  \\
	DMU (Regression) & \textbf{25.27}  &  21.23 & \textcolor[RGB]{0,200,0}{12.15}  & \textcolor[RGB]{200,0,0}{\textbf{1.21}}  \\
	\bottomrule
\end{tabular}
}
\vspace{-8pt}
\caption{Different regularizers in  30\%data$\rightarrow$100\%data on GLDv2-test. The backbone is ResNet50.}
\label{tab:exp-diff-compatible-loss}
\vspace{-8pt}
\end{table}

\subsubsection{Effectiveness of Forward Adaptation.}
Results in the third row of Table~\ref{tab:exp-ablation} shows that mere forward-adaptive head can extremely boost the new-to-old retrieval performance compared with baseline. Evolving old features (especially poor features) towards to the new embedding space contributes to the improvement.

We ablate the hidden dimension of the forward-adaptive head $\psi$
and report results in Table~\ref{tab:exp-fa-dimension}.
Considering the trade-off between the accuracy and the refresh efficiency, we adopt the median hidden dimension 1024 as the default setting in the following experiments unless stated otherwise.

The MACs (Multiply-Accumulate Operations) of a Forward-adaptive head with 1024-dim is about 8M, which is marginal to refresh gallery database. Specifically, refreshing a billion features with FA head with 1024 (running with TensorRT on NVIDIA Tesla T4 GPU) only consumes less than 8 hours. At a negligible cost, user experience can be gradually improved with DMU framework.

\subsubsection{Different Compatible Regularizers.}
To elaborate our proposed selective backward compatibility is independent with compatible regularizers, we apply our DMU framework on another regression version~\cite{budnik2020asymmetric} which minimizes the cosine embedding distance between new and old features.
\begin{equation} \label{eq:cosine-embedding-loss}
  \ell_{\rm reg}(x;\phi_{\rm n}, \phi_{\rm o}) = 1 - \cos\langle\phi_{\rm n}(x), \phi_{\rm o}(x)\rangle.
\end{equation}

We replace $\ell_{\rm arc}(x;\phi_{\rm n}, \phi_{\rm o})$ with $\ell_{\rm reg}(x;\phi_{\rm n}, \phi_{\rm o})$ in Equation~\ref{eq:sbc_loss} and \ref{eq:fa-loss}. Table~\ref{tab:exp-diff-compatible-loss} displays the comparison using different types of compatible regularizers. We observe that our method can also alleviate the new-to-new degradation, which indicates dynamically inheriting discriminative old features is effective. To summarize, our proposed paradigm, Darwinian Model Upgrades, is 
general and effective.

\subsubsection{Sequential Model Upgrades.}
To further explore the advantage of DMU, we conduct sequential model upgrades experiments and the results are shown in Table~\ref{tab:exp-sequential-upgrades}. As can be seen, our proposed DMU paradigm consistently surpasses the previous BCT paradigm in both new-to-old and new-to-new retrieval performance. It demonstrates again that our novel selective backward compatibility loss can facilitate the better query features and the introduced forward-adaptation head can achieve the better gallery features.

\subsubsection{Comparison with Other Methods.}
We compare with different SoTA compatible learning works in Table~\ref{tab:exp-Comparison-sotas}.
When compared with the Focal Distillation method, our DMU can achieve comparable performance in new-to-new metric and surpass it largely in new-to-old which exactly demonstrates the superiority of our re-weight scheme than the Focal Distillation one.
Besides, when compared with UniBCT, DMU achieves both better new-to-old and new-to-new performances. It demonstrates that our selective backward-compatible loss works better in noisy samples resistance than the graph transition proposed in UniBCT.

\begin{table}[t]
\renewcommand\arraystretch{1.3}
\centering
\setlength{\tabcolsep}{0.1mm}{	
\begin{tabular}{lcccc}
	\toprule
	\multirow{2}[3]{*}{Dimension} & \multicolumn{4}{c}{GLDv2-test} \\
	\cmidrule(r){2-5}
	~ & ${\cal M}(\phi_{\rm n},\phi_{\rm n})$ & ${\cal M}(\phi_{\rm n},\phi_{\rm o})$ & \quad\textcolor[RGB]{0,200,0}{$\Delta_{\uparrow}$}\quad & \quad\textcolor[RGB]{200,0,0}{$\Delta_{\downarrow}$}\quad \\
	\midrule
	BCT(CVPR'20) &  24.55 & 21.12  &  \textcolor[RGB]{0,200,0}{11.57} & \textcolor[RGB]{200,0,0}{4.03}   \\
	FD(CVPR'21) &  \textbf{25.24}  &  20.02  &  \textcolor[RGB]{0,200,0}{5.76} & \textcolor[RGB]{200,0,0}{\textbf{1.3}} \\
	UniBCT(IJCAI'22) & 24.60 &  21.77 & \textcolor[RGB]{0,200,0}{15.00}  & \textcolor[RGB]{200,0,0}{3.83}  \\
	ours & 25.00 & \textbf{22.56}  & \textcolor[RGB]{0,200,0}{\textbf{19.18}}  & \textcolor[RGB]{200,0,0}{2.27}  \\

	\bottomrule
\end{tabular}
}
\vspace{-8pt}
\caption{Comparison with SOTAs in  30\%data$\rightarrow$100\%data on GLDv2-test. The backbone is ResNet50.}
\label{tab:exp-Comparison-sotas}
\vspace{-8pt}
\end{table}

\section{Conclusions}
Due to the undifferentiated compatibility constraints, existing backward-compatible works are stuck in dilemma between new feature discriminativeness and new-to-old compatibility. To tackle the issue, we introduce a novel paradigm, Darwinian Model Upgrade  with selective compatibility, which tends to inherit discriminative old features with selective backward compatible training and evolves old features to become increasingly well-suited to the better latent space with a light-weight forward-adaptive head.
With DMU, the query embedding quality is improved by the new backbone in the meanwhile the gallery is quickly refreshed by the forward-adaptive head, rendering a continual cross-model retrieval boost.
We demonstrate the effectiveness of DMU in many mainstreaming datasets including Google Landmark, ROxford, RParis, MS1Mv3 and IJB-C. 
We hope our work can inspire the community to discover proper model upgrading approaches.

\bibliography{aaai23}

\newpage
\clearpage

\section*{Appendix A}
In this section, we will describe more experimental details and the pseudo code of Darwinian Model Upgrades for reproductive results, and the source code will be released upon this paper is accepted.

\subsection*{Data Allocation.}
To simulate real-world compatible scenarios, we adopt three types of data allocation following~\cite{zhang2022towards}: (1) in extended-data (30\%data$\rightarrow$100\%data), the old training set is composed with 30\% data randomly selected from the whole dataset (GLDv2-train or MS1Mv3), and the new training set contains all data; (2) in open-data (30\%data$\rightarrow$70\%data), the old training set and the new one has no overlap but shares the same classes; (3) in extended-class (30\%data$\rightarrow$100\%data), the old training set consists of 30\% classes randomly picked from the whole dataset, and the new one contains the rest of classes. More details are illustrated in Table~\ref{tab:data-allocation-landmark} and~\ref{tab:data-allocation-face}.

\begin{table}[h]
\renewcommand\arraystretch{1.3}
\centering
\setlength{\tabcolsep}{2mm}{	
\begin{tabular}{lccc}
            \toprule
            ~ & Usage &  Images &  Classes \\
            \midrule
            \multirow{3}{*}{\makecell[c]{ {Data-split} }} & 30\% & 445,419 & 81,313 \\
            ~ & 70\% & 1,135,051 & 81,313 \\
            ~ & 100\% & 1,580,470 & 81,313\\
            \cmidrule(r){1-4}
            \multirow{2}{*}{\makecell[c]{ {Class-split} }} & 30\% & 389,526 & 20,328 \\
            ~ & 100\% &  1,580,470 & 81,313 \\
            \bottomrule
        \end{tabular}
}
\caption{Data allocation of GLDv2-train dataset.}
\label{tab:data-allocation-landmark}
\end{table}

\begin{table}[h]
\renewcommand\arraystretch{1.3}
\centering
\setlength{\tabcolsep}{2mm}{	
\begin{tabular}{lccc}
            \toprule
            ~ & Usage &  Images &  Classes \\
            \midrule
            \multirow{3}{*}{\makecell[c]{ {Data-split} }} & 30\% & 1,511,514 & 93,431 \\
            ~ & 70\% & 3,667,996 & 93,431 \\
            ~ & 100\% & 5,179,510 & 93,431\\
            \bottomrule
        \end{tabular}
}
\caption{Data allocation of MS1M-v3 dataset.}
\label{tab:data-allocation-face}
\end{table}

\subsection*{Pseudo Code of Darwinian Model Upgrades.}
The pseudo code of our manner is illustrated in Algorithm~\ref{alg:code_dmu}.

\begin{algorithm*}[t]

\algcomment{\fontsize{10pt}{0em}\selectfont \texttt{bmm}: batch matrix multiplication; \texttt{mm}: matrix multiplication; \texttt{cat}: concatenation; \texttt{t()}: matrix transpose.
}
\definecolor{codeblue}{rgb}{0.25,0.5,0.5}
\lstset{
  backgroundcolor=\color{white},
  basicstyle=\fontsize{10pt}{10pt}\ttfamily\selectfont,
  columns=fullflexible,
  breaklines=true,
  captionpos=b,
  commentstyle=\fontsize{10pt}{10pt}\color{codeblue},
  keywordstyle=\fontsize{10pt}{10pt},
}
\begin{lstlisting}[language=python,escapeinside={(*}{*)}]
# old_model: pretrained and fixed old encoder, old_model.fc is the old classifier (no gradient)
# new_model: new encoder, new_model.fc is the classifier
# psi: forward-adaptive head
# N: batch size, D: feature dimension

for (x, targets) in loader:  # load a mini-batch x with N samples
    x_old = aug(x)  # a randomly augmented version
    x_new = aug(x)  # another randomly augmented version
    
    with torch.no_grad():
        old_feat = old_model.forward(x_old).detach()  # tensor shape: NxD
    fa_feat = psi.forward(old_feat)
    new_feat = new_model.forward(x_new)
    
    # Classification loss, Eq.(3)
    cls_logits = new_model.fc(new_feat)
    cls_loss = ArcFaceLoss(cls_logits, targets)
    
    # Selective backward compatible loss, Eq.(9) and (10)
    old_cls_logits = softmax(old_model.fc(old_feat))
    Lambda = -torch.sum(torch.mul(old_cls_logits, torch.log(old_cls_logits)), dim=1) # Eq.(9)
    dynamic_weight = torch.div(1-softmax(Lambda, dim=0), N-1) # Eq.(10)
    sbc_loss = ArcFaceLoss(cls_logits, targets, reduction='none')
    sbc_loss = torch.sum(torch.mul(sbc_loss, dynamic_weight))
    
    
    # Forward-adaptive loss
    fa_cls_logit = new_model.fc(fa_feat)
    fa_cls_loss = ArcFaceLoss(fa_cls_logit, targets)
  
    # SGD update: new model and forward-adaptive head
    loss = cls_loss + sbc_loss + fa_cls_loss
    loss.backward()
    update(new_model.params)
    update(psi.params)
\end{lstlisting}
\caption{Pseudocode of Darwinian Model Upgrades in a PyTorch-like style.}
\label{alg:code_dmu}
\end{algorithm*}

\newpage
\section*{Appendix B}
In this section, more experimental results are provided to further demonstrate the effectiveness of our method.

\subsection*{Model Compatibility between Transformer and CNN.}
To validate the generalization of Darwinian Model Upgrades, we conduct compatible training between different architectures (\ie, Swin-Transformer-tiny~\cite{liu2021swin} and ResNet50~\cite{he2016deep}). As shown in Table~\ref{tab:exp-landmark-2}, DMU significantly surpasses the baseline in both new-to-new and new-to-old protocols.

\begin{table}[h]
\renewcommand\arraystretch{1.3}
\centering
\setlength{\tabcolsep}{0.1mm}{	
\begin{tabular}{llcccc}
	\toprule
	\multirow{2}[3]{*}{Scenario} & \multirow{2}[3]{*}{Method} & \multicolumn{4}{c}{GLDv2-test} \\
	\cmidrule(r){3-6}
	~ & ~ & ${\cal M}(\phi_{\rm n},\phi_{\rm n})$ & ${\cal M}(\phi_{\rm n},\phi_{\rm o})$ & \quad\textcolor[RGB]{0,200,0}{$\Delta_{\uparrow}$}\quad & \quad\textcolor[RGB]{200,0,0}{$\Delta_{\downarrow}$}\quad \\
	\midrule

	\multirow{4}{*}{\makecell[l]{30\%data\\$\rightarrow$ \\ 100\%data}} &
	$\phi_{\rm o}^{r50}$ &  18.93 &  - &  - & -  \\
	~ & $\phi_{\rm n}^{swin}$ (oracle)& 26.80  &  - &  - & - \\
	~ & $\phi_{\rm n}^{swin}$ (BCT) & 25.53 & 21.62  &  \textcolor[RGB]{0,200,0}{14.21}  & \textcolor[RGB]{200,0,0}{4.74} \\
	~ & $\phi_{\rm n}^{swin}$ (ours) & \textbf{25.89} & \textbf{22.47} &  \textcolor[RGB]{0,200,0}{\textbf{18.70}}  & \textcolor[RGB]{200,0,0}{\textbf{1.34}} \\	
	\hline
	\bottomrule
\end{tabular}
}
\caption{Comparison of baselines (BCT) and our method (DMU) on  GLDv2-test. All models are trained on GLDv2-train dataset. The evaluation protocols are mAP ($\mathcal{M}(\cdot,\cdot)$, larger is better), upgrade gain ($\Delta_{\uparrow}$, larger is better) and discriminativeness degradation ($\Delta_{\downarrow}$, smaller is better). The architectures are ResNet50 (r50) and SwinTransformerTiny (swin). A negative value for degradation means that the degradation has been fully resolved, and the new model even benefits from backward-compatible regularization.}
\label{tab:exp-landmark-2}
\vspace{-18pt}
\end{table}

\end{document}